\documentclass[conference]{IEEEtran}
\IEEEoverridecommandlockouts
\usepackage{cite}
\usepackage{amsmath,amssymb,amsfonts}
\usepackage{algorithmic}
\usepackage{graphicx}
\usepackage{textcomp}
\usepackage{xcolor}
\usepackage{subcaption}
\usepackage{booktabs, adjustbox}
\usepackage{flushend}

\def\BibTeX{{\rm B\kern-.05em{\sc i\kern-.025em b}\kern-.08em
    T\kern-.1667em\lower.7ex\hbox{E}\kern-.125emX}}

\begin{document}

\newtheorem{remark}{Remark}

\title{\bf \Large Towards Explainability in Modular Autonomous Vehicle Software
}

\author{Hongrui Zheng$^{*}$, Zirui Zang$^{*}$, Shuo Yang$^{*}$, Rahul Mangharam
\thanks{$^*$Authors contributed equally. All authors are with University of Pennsylvania, Department of Electrical and Systems Engineering, 19104, Philadelphia, PA, USA. Emails: \{\tt\small hongruiz, zzang, yangs1, rahulm\}@seas.upenn.edu}
}

\maketitle

\begin{abstract}
Safety-critical Autonomous Systems require trustworthy and transparent decision-making process to be deployable in the real world. The advancement of Machine Learning introduces high performance but largely through black-box algorithms.
We focus the discussion of explainability specifically with Autonomous Vehicles (AVs). As a safety-critical system, AVs provide the unique opportunity to utilize cutting-edge Machine Learning techniques while requiring transparency in decision making. Interpretability in every action the AV takes becomes crucial in post-hoc analysis where blame assignment might be necessary.
In this paper, we provide positioning on how researchers could consider incorporating explainability and interpretability into design and optimization of separate Autonomous Vehicle modules including \textit{Perception, Planning, }and \textit{Control}.
\end{abstract}
\section{Introduction}
According to the Morning Consult and Politico poll \cite{mc-politico-poll}, only 16\% of respondents are “very likely” to ride as a passenger in an autonomous vehicle, while 28\% of respondents state that they “not likely at all”. Moreover, only 22\% of respondents believe self-driving cars are safer than the average human driver, while 35\% of them believing self-driving cars are less safe than the average human driver. The public's distrust in Autonomous Vehicles (AV) shows that improving explainability in AV software is a necessity. 

There exist many surveys on explainable AI (XAI) and robotics \cite{dovsilovic2018explainable, islam2021explainable, tjoa2020survey, sakai2022explainable}. Specifically, \cite{zablocki2022explainability, atakishiyev2021explainable ,omeiza2021explanations} surveys explainability in Autonomous Driving. Atakishiyev et al. \cite{atakishiyev2021explainable} believes that AVs need to provide regulatory compliant operational safety and explainability in real-time decisions. It focuses on providing discussion through the cause-effect-solution perspective. Zablocki et al. \cite{zablocki2022explainability} provides an in-depth overview of XAI methods in deep vision-based methods, but is limited to the scope of perception only. Omeiza et al. \cite{omeiza2021explanations} also provides an overview of explanations in AVs in the full self-driving pipeline.
Gilpin et al. \cite{gilpin2018explaining} proposes explainability as the trade-off between \textit{interpretability} and \textit{completeness}. As described in \cite{gilpin2018explaining}, to be interpretable is to describe the internals of a system in such a way that is understandable to humans; to be complete is to describe the operation of a system in an accurate way. 

We position ourselves to provide insight in augmenting explanability in Autonomous Vehicle's sense-plan-act software modules as a task of balancing interpretability and completeness.
In this paper, we look at the explainability in existing works and in our recent contributions in localization, planning, and control. In each case, we want to be able to quantify the uncertainty at each step of the decision making and interpret the provenance of the outcome of the algorithm.

\section{Explainability in Localization}

Robot localization is a problem of finding a robot’s pose using a map and sensor measurements, such as LiDAR scans. The map is pre-built and the environment is assumed to not change significantly after the map is captured. It is crucial for any moving robot to interact with the physical world correctly. However, the problem of finding the mappings between measurements and poses can be ambiguous, because sensor measurements from multiple distant poses can be similar. Therefore, to tightly integrate the localization module with other parts of the software stack and for the engineers implementing and tuning the algorithm, the explainability of localization algorithms using neural networks becomes important. We need to estimate the uncertainty of the localization results, and in the worst case, to know when and why the robot fails to localize on a certain map. 

Monte Carlo Localization (MCL)\cite{dellaert1999monte}, the widely adopted method, uses random hypothesis sampling and sensor measurement updates to infer the pose. In MCL, the proposed particles are explicit poses on the map and we can interpret the distribution of the particles as the uncertainties. The random generation of the particles can be tuned with parameters that have physical meaning, providing an interface for humans to adjust the behavior of the algorithm. Many developments in localization seek to improve within the framework of MCL.\cite{zhang2018robust,chen2021icra,chen2020learning} Although particle filter has been a popular localization framework for its robustness and reliable performance, it introduces random jitter into the localization results.

Other common approaches are to use Bayesian filtering\cite{fox2003bayesian} or to find  more distinguishable global descriptors on the map\cite{dube2020segmap,sarlin2019coarse}. In Bayesian filtering, the explainability lies in the conditional probability attached with the motion model and each measurement. The estimation of such probability is challenging. For the global descriptor approach, oftentimes manual selection of map features are needed, which increases the explainability of the system, but also increases the human workload and reduces robustness. Developments in localization research usually propose better measurement models or feature extractors within these frameworks. \cite{uy2018pointnetvlad,sun2020localising}. 

Recent research in localization has also focused on the use of learning-based methods outside of the above frameworks \cite{lu2019l3}. Although learning-based methods may provide better localization precision with lower latency, the interpretability of the method decreases. While the traditional localization methods can be manually tuned according to the specific user scenarios, learning-based localization methods are usually not tunable once the network is trained. Uncertainty estimations of the neural networks also become a challenge for learning-based methods. There are efforts to approximate the uncertainty\cite{cai2019hybrid,kendall2016modelling,deng2022deep}, but it hasn't been widely applied.

\textbf{Our contribution:} In our recent paper, Local\_INN, we proposed a new approach to frame the localization problem as an ambiguous inverse problem and solve it with an invertible neural network (INN) \cite{zang2022local_inn}. It stores the map data implicitly inside the neural network. With the assumption that the environment doesn't not change significantly from the map, by evaluating the reverse path of the neural network, we can get robot poses from LiDAR scans. It also provides uncertainty estimation from the neural network and is capable of learning and providing localization for complex environments.

Localization is an inverse problem of finding a robot’s pose using a map and sensor measurements. This reverse process of inferring the pose from sensor measurements is ambiguous. Invertible neural networks such as normalizing flows\cite{papamakarios2021normalizing} have been used to solve ambiguous inverse problems in various fields\cite{ardizzone2018analyzing,ardizzone2019guided,adler2019uncertainty,WehRud2021}. The version of normalizing flows we used is called RealNVP\cite{dinh2016density}, which uses a mathematical structure called coupling layers to ensure the invertibility while performing transformations with arbitrary neural network layers, such as MLPs. This framework of solving inverse problems with normalizing flows was introduced by Ardizonne et al.\cite{ardizzone2018analyzing} and was later extended by \cite{ardizzone2019guided,winkler2019learning} to include a conditional input that is concatenated to the vectors inside the coupling layers. They proposed to use normalizing flows to learn a bijective mapping between two distributions and use a normal-distributed latent variable to encode the lost information in training due to the ambiguity of the problem. The network can be evaluated in both forward and reverse paths. During the evaluation, repeatedly sampling the latent variable can give the full posterior distribution given the input. 

\begin{figure}[t]
\centering
\vspace{-10pt}
\includegraphics[width=0.98\columnwidth]{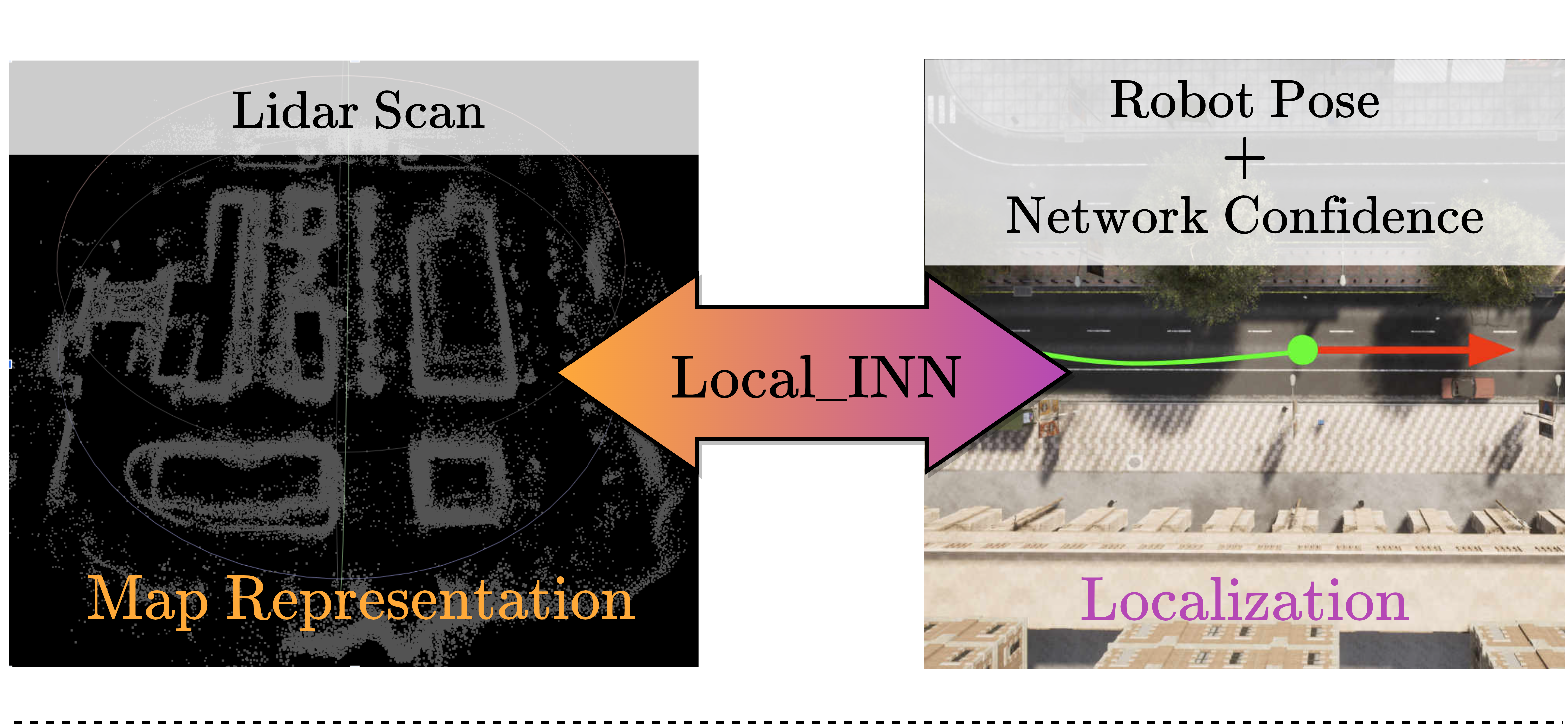}
\includegraphics[width=0.98\columnwidth]{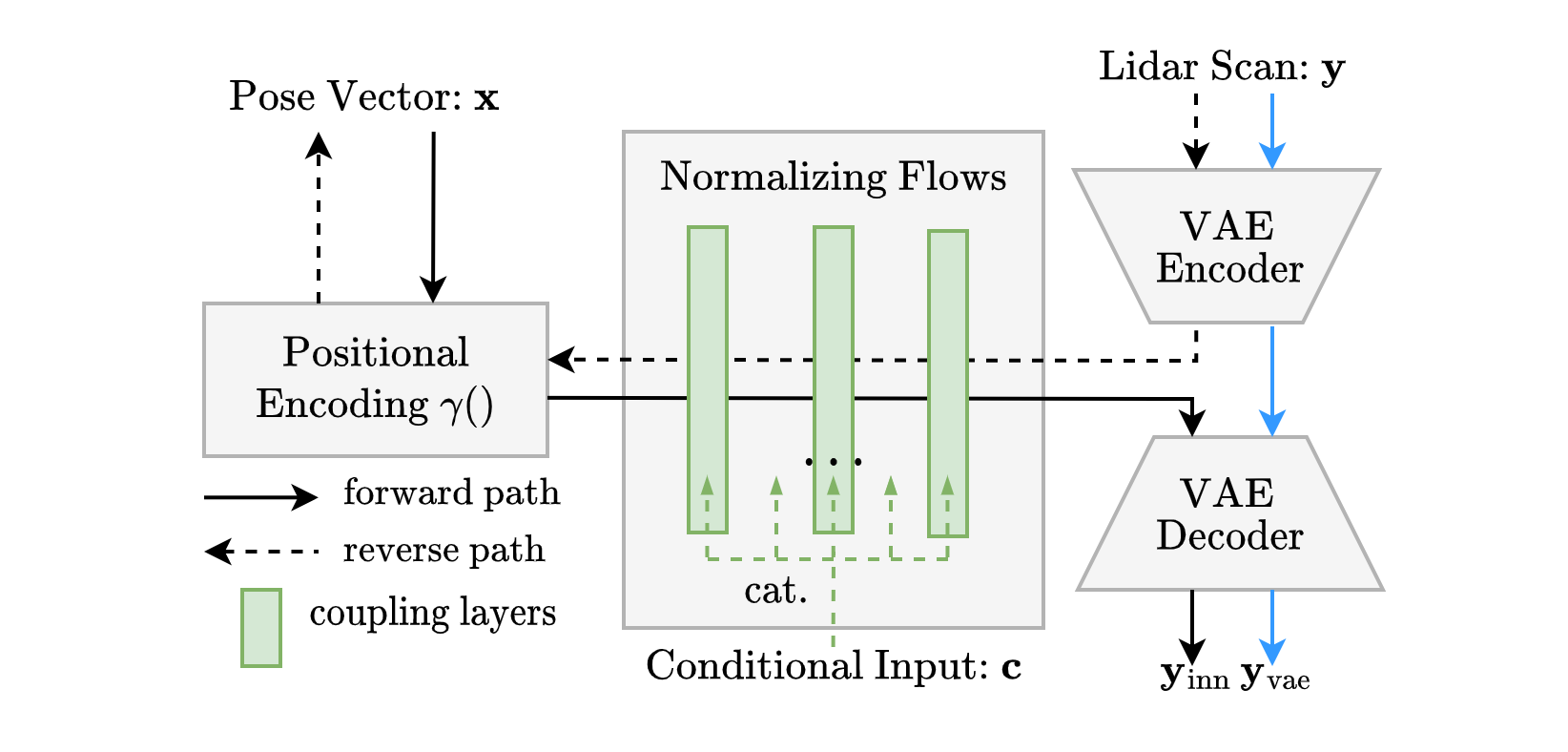}
\vspace{-10pt}
\caption{\small\textbf{Top}: Evaluation of Local\_INN in forward direction gives compressed map information, and in the reverse direction gives accurate localization with fast runtime and uncertainty estimation. \textbf{Bottom}: Network structure of Local\_INN. Solid arrows are from pose to lidar scan. Dashed arrows are from lidar scan to pose. Conditional input is calculated from the previous pose of the robot.}
\label{fig:fig_local_inn}
\vspace{-15pt}
\end{figure}

In Local\_INN, we use pose-scan data pairs to train such a bijective mapping. As shown in Fig. \ref{fig:fig_local_inn}, The forward path is from pose to scan and the reverse path is from scan to pose. We use a conditional input calculated from the previous pose of the robot to reduce the ambiguity of the problem. Because INNs require the same input and output dimensions, we use a Variational Autoencoder\cite{kingma2013auto} to reduce the dimension of the LiDAR scans and use Positional Encoding\cite{vaswani2017attention} to augment that of the poses. The network is trained with supervised loss functions on both sides. The data used for training the Local\_INN can be simulated or real data recorded from LiDARs. In our experiments, we tested on both real and simulated data with 2D and 3D LiDARs. To collect training data, we uniformly sample $x, y$ position and heading $\theta$ on the drivable surface of each map, and use a LiDAR simulator to find the corresponding LiDAR ranges. This means the trained network will be able to localize everywhere on the map. For each different map, we need to train a separate network. Map files are compressed inside the neural network and are no longer needed during evaluation. 

\begin{table}[b]
\centering
\vspace{-15pt}
\setlength{\tabcolsep}{2pt}
\caption{Local\_INN Experiments: Map Reconstruction and RMS Localization Errors with 2D LiDAR ($xy$[m], $\theta$[$^\circ$])}
\begin{tabular}{ l c c }

\toprule 
\multicolumn{1}{c}{} & 
\multicolumn{1}{c}{Race Track (Simulation)} & 
\multicolumn{1}{c}{Outdoor (Real)}  \\ 
\midrule

\begin{tabular}[c]{@{}c@{}}Original Map \\ \textcolor{orange}{Reconstruction} \\
\textcolor{violet}{Test Trajectory}
\end{tabular}
& \multicolumn{1}{c}{\adjincludegraphics[valign=c,width=0.43\columnwidth]{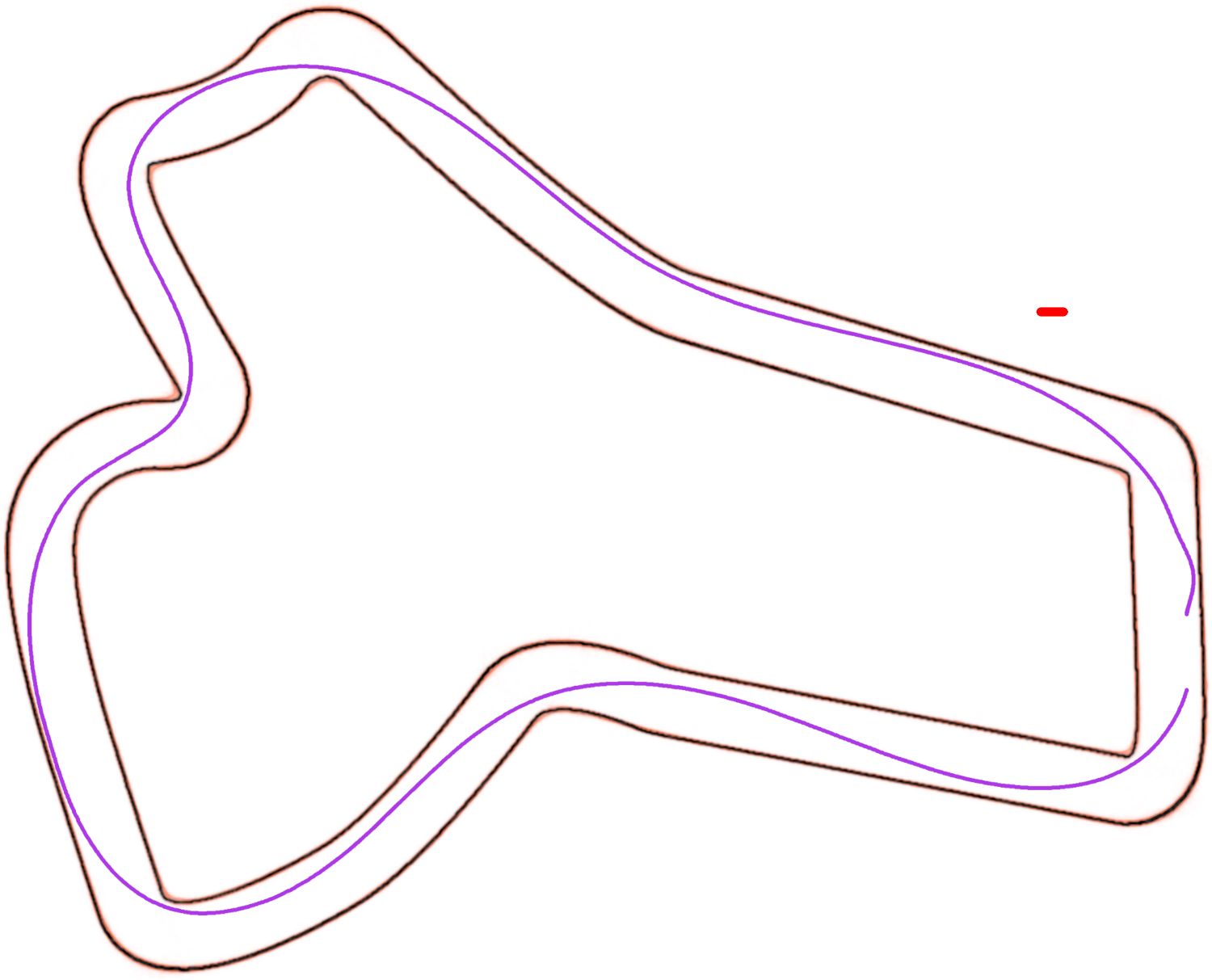} }
& \multicolumn{1}{c}{\adjincludegraphics[valign=c,width=0.30\columnwidth]{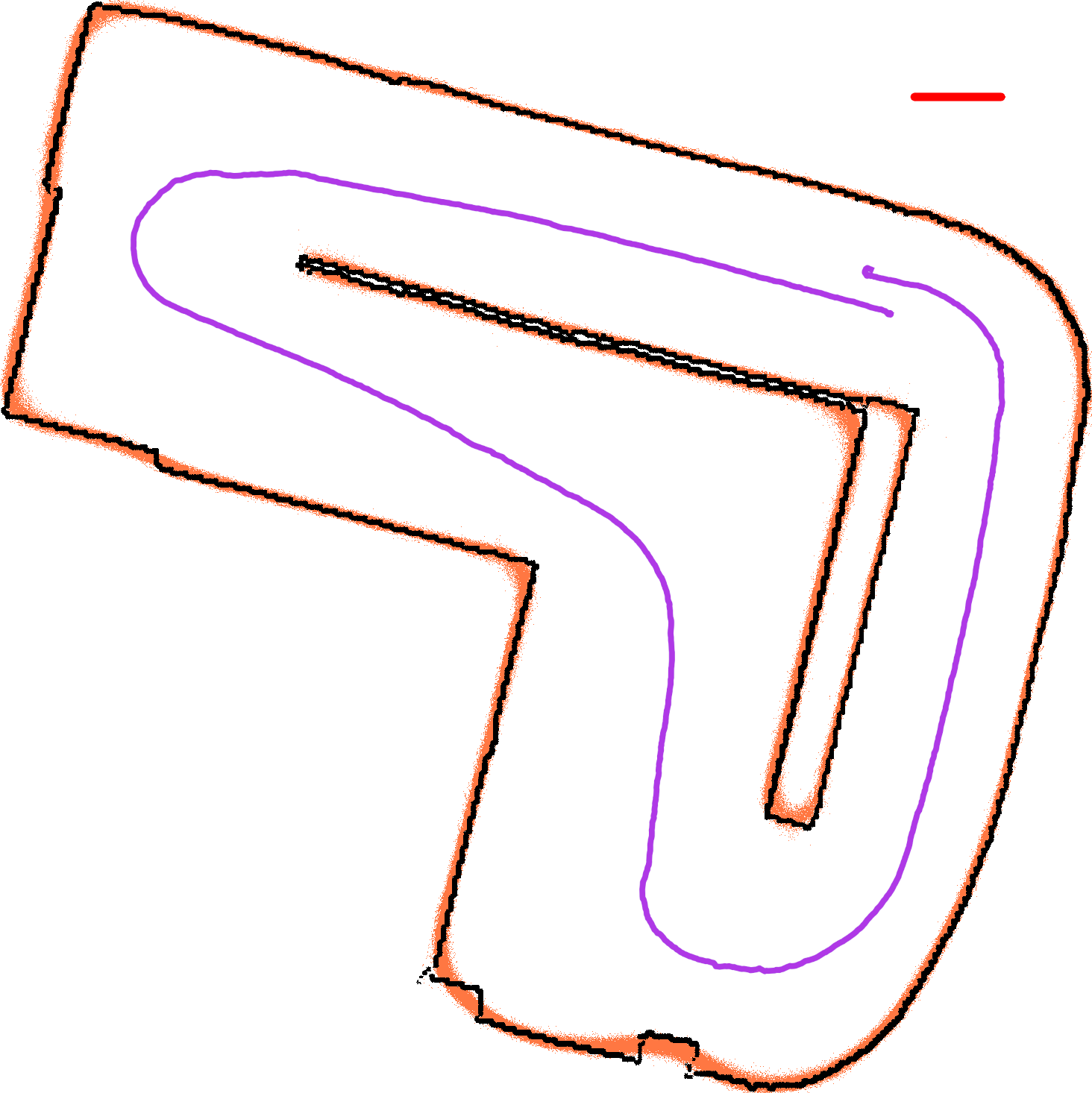} }
\\ \midrule

Online PF
& $0.168, 2.107$
& $0.047, 1.371$
\\
Local\_INN+EKF
& $\mathbf{0.056}, \mathbf{0.284}$
& $\mathbf{0.046}, \mathbf{1.130}$

\\ \bottomrule

\end{tabular}
\label{table_2d}
\end{table}

We claim that INN is naturally suitable for the localization problem with improved explainability compared to other learning-based methods. In particular, uncertainty estimation and map representation are the two advantages that Local\_INN provides in the context of explainability.

\begin{figure*}[t]
\centering
\includegraphics[width=1.70\columnwidth]{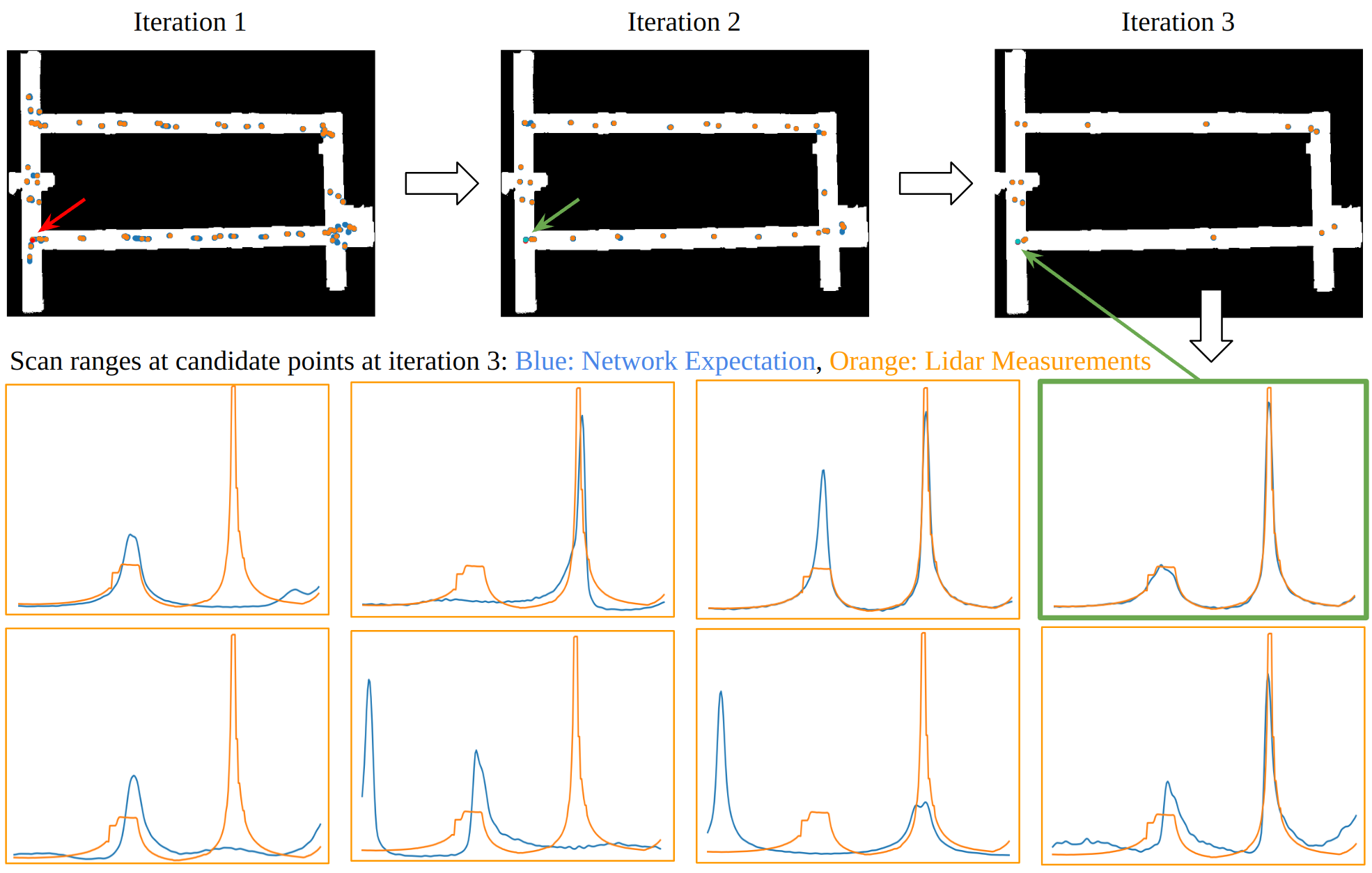}
\caption{Example of global localization finding the correct pose at the 2nd iterations (green arrow). \textbf{Top}: Narrowing down of the candidate poses in the first 3 iterations. We can see candidate poses on the map (orange dots), correct pose (red dot), selected pose (green dot). \textbf{Bottom}: Examples of LiDAR scan ranges at candidate poses at iteration 3 (orange boxes), and at the selected pose (green box). In the range plots, the horizontal axis is the angle of LiDAR scans and vertical axis is the measured distance. We can see a comparison of network expected ranges (blue curve) at various poses, and the actual LiDAR measurement (orange curve). The correct pose is selected where the measurement best matches the expected shape. The network is trained on simulated data and tested on real LiDAR data.}
\label{fig_local_inn_global}
\vspace{-10pt}
\end{figure*}

\subsection{Explainability from Uncertainty Estimation} 

When we use Local\_INN to localize, the input to the reverse path of the INN consists of the LiDAR scans concatenated with a latent vector that is sampled from normal distribution. With this sampling of latent vector, the network can output not just a pose but a distribution of inferred poses. The covariance of this distribution can be used as the confidence of the neural network when fusing with other sensors. Uncertainty estimation improves explainability by providing information on the measurement quality of the prediction. Compared to learning methods that do not provide uncertainty estimates, it is much easier to determine whether the prediction of the neural network is lower in accuracy due to higher uncertainty, and improve the prediction results by augmentation. In our experiments, we used an EKF to fuse the localization result with the odometry information. The results show that this fusion significantly improved localization accuracy where the map geometry is ambiguous, which means this covariance is very effective in revealing the confidence of the network. 

As shown in Table I, The accuracy of Local\_INN is at par with the current localization efforts. See \cite{zang2022local_inn} for a comparative analysis of Local\_INN localization accuracy in 2D and 3D maps.

\subsection{Explainability from Map Representation}

Local\_INN provides an implicit map representation and a localization method within one neural network. The guaranteed invertibility of the neural network provides the use a direct way to check the neural network's 'understanding' of the map by reproducing part of the map with poses. That is, we can compare the reconstructed map to the original map to see how much detail is used by the neural network in localization. Again, this feature improves explainability in failure scenarios. When the localization fails, this comparison can help us explain the failure and guide us in improving the methods. For example, we can train with more data from a particular location on the map that was difficult to localize in.

As an example of how the stored map information in the forward path of the neural network can help us explain the localization results, let us consider the algorithm for global localization. Global localization is needed when a robot starts with an unknown pose or when the robot encounters the 'kidnapping problem'. In this case, it is challenging to find the correct position on the map due to the ambiguity of problem. MCL algorithms usually do global localization by spreading the covariance all around the map and using iterations of control inputs and measurements to decrease the covariance, which gives an explicit visualization to see the progress of the global localization processes. For other learning-based method, this process is usually hard to explain as we rely on the neural network to output poses as a black box.

With Local\_INN, we can randomly initialize a set of random poses on the map as conditional inputs and use new lidar scans to narrow down the assumptions. In other words, we initially have multiple random assumptions of the robot's location on the map and use them as the conditional inputs for the nerual network. As shown in figure \ref{fig_local_inn_global} iteration 1, when we input a LiDAR scan to the network along with these assumptions, it will output multiple possible pose distributions. In our algorithm, for each possible pose distribution, we compare the sensor measurement with what the neural network expects at this location, and use the reciprocal of the error term to weight the assumptions differently. The weights for the assumptions is used to determine the amount of latent variable we use. This process repeats with each new LiDAR scan we get from the sensor. In our experiments, the convergence of candidate poses is fast and accurate. As shown in iteration 2 and 3 in figure \ref{fig_local_inn_global}, even if we still track multiple poses, in this example, the correct pose is determined at the 2nd iteration in a highly symmetrical map. The low part of figure \ref{fig_local_inn_global} shows plots of the expected lidar scan from the neural network and the current LiDAR measurement. This reveals the black-box process of the neural network so that we can see why multiple poses are possible and how we should decide which one to pick.

To summarize, the explainability of localization methods generally lies in the uncertainty estimation and the ability to explain and tune the methods when localization fails. Traditional localization methods usually offer higher interpretability than learning-based methods, whereas the learning-based methods can provide better empirical performance. The new method, Local\_INN, we recently proposed uses an invertible network network architecture to solve the localization problem. It offers interpretability by giving uncertainty estimation through the covariance of the inferred pose distributions, and by ensuring the invertibility of the network so that we can reveal the what information the neural network is using during the localization. At the same time, it does sacrifice completeness by using a Variational Autoencoder (VAE) to model a latent space of the LiDAR scans.
\section{Explainability in Planning}
Planning is the task of finding a viable motion plan for the robot to reach some predetermined goal given the current observation of the environment through various sensors. The planning step is usually the next step after perception, or localization. Traditionally, sampling-based and model-predictive methods are the most popular choices in Autonomous Vehicle motion planning. Planning algorithms provide explanability through human-designed objectives: e.g. maximizing the distance of a planned trajectory to obstacles and undrivable areas, maximizing velocity on the trajectories, minimizing the lateral acceleration the vehicle experiences on a planned trajectory. We propose a unique position in adding interpretability and completeness to planning algorithms: explainability through abstraction. Next, we show our approach based on our recent paper on Game-Theoretic objective space planning \cite{zheng2022game}.

\begin{figure}
    \centering
    \includegraphics[width=1\columnwidth]{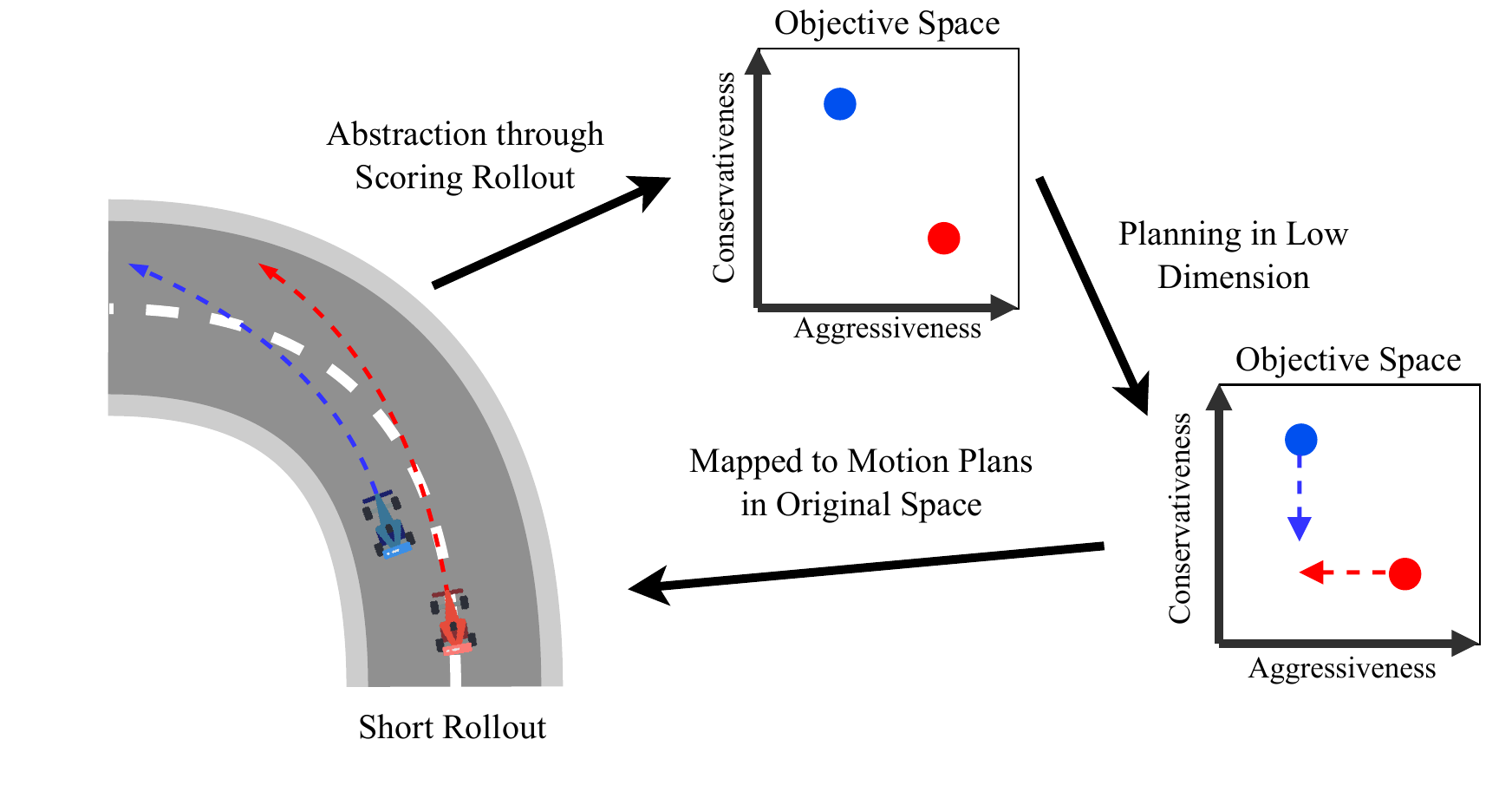}
    \vspace{-10pt}
    \caption{Overview of Game-theoretic Objective Space Planning}
    \label{fig:obj_planning}
    \vspace{-10pt}
\end{figure}

In this case study, our primary context is two-player racing games in close proximity. An overall depiction of the pipeline is shown in Figure \ref{fig:obj_planning}. We choose a game theoretic approach that models racing as a \textit{zero-sum extensive game} with \textit{imperfect information} and \textit{perfect recall}. Extensive games model sequential decision-making of players and naturally form a game tree where each node is a decision point for a player. However, the planning problem presented in autonomous racing is continuous, and the state space of the agents, in turn, the game tree in the extensive game, will also be infinitely large if we model the game in the vehicle's planning space. Since the decision made by a game-theoretic algorithm in the planning space cannot be explained in a way that a human can understand, we use a lower dimensional space for planning. We define the notion of \textit{Objective Space} $\mathcal{O}$. For each short rollout in an adversarial environment, we can compute multiple metrics regarding this agent's performance, such as safety and aggressiveness. These metrics also add to the interpretability of our planning algorithm while not losing the completeness.
$\mathcal{O}$ models the average outcome of each agent against competent opponents. Using $\mathcal{O}$, our planner maps complex agent behaviors to a lower dimension where only the episodic outcome is recorded instead of the entire decision-making process in the planning space. 
We define an action in our game as movements in a dimension of $\mathcal{O}$. This action space is analogous to the planning space in a grid world with actions that move the agent to a neighboring cell. This means the planning problem is much simpler than the original problem.

In our case study, we choose \textit{aggressiveness} and \textit{restraint} as the two dimensions of $\mathcal{O}$. Aggressiveness is scored on an agent's lead over the other at the end of the rollout, and restraint is scored on an agent's time to collision to the environment and the other agent.
Two movements are available for each dimension: increasing or decreasing for a fixed distance along the axis.
For example, four discrete actions are available at each turn when $\mathcal{O}\in\mathbb{R}^2$. Even with the formulation of agent action space, the possible objective values of an opponent agent or possible position in $\mathcal{O}$ is still infinite. We propose a predictive model for regret values within Counterfactual Regret Minimization (CFR) \cite{zinkevich_regret_2007} to make the problem tractable. Finally with head-to-head racing experiments, we demonstrate that using the proposed planning pipeline above significantly improves the win rate that generalizes to unseen opponents in an unseen environment.

\subsection{Explainability in Agent Actions}
\begin{figure}
    \centering
    \includegraphics[width=1\columnwidth]{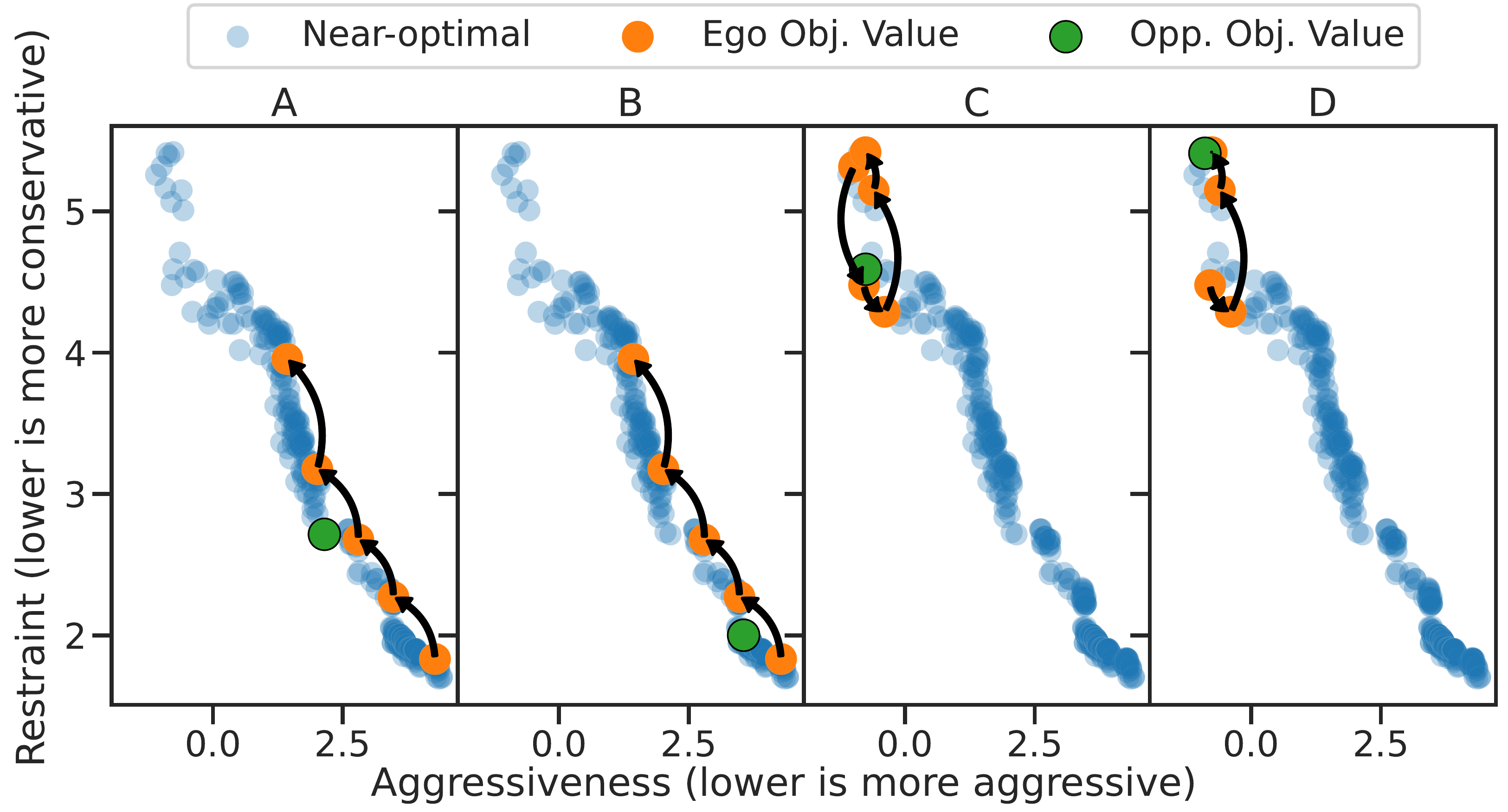}
    \caption{Trajectories of ego moves in $\mathcal{O}$.}
    \label{fig:cases_all}
\end{figure}
\begin{figure}
    \centering
    \includegraphics[width=1\columnwidth]{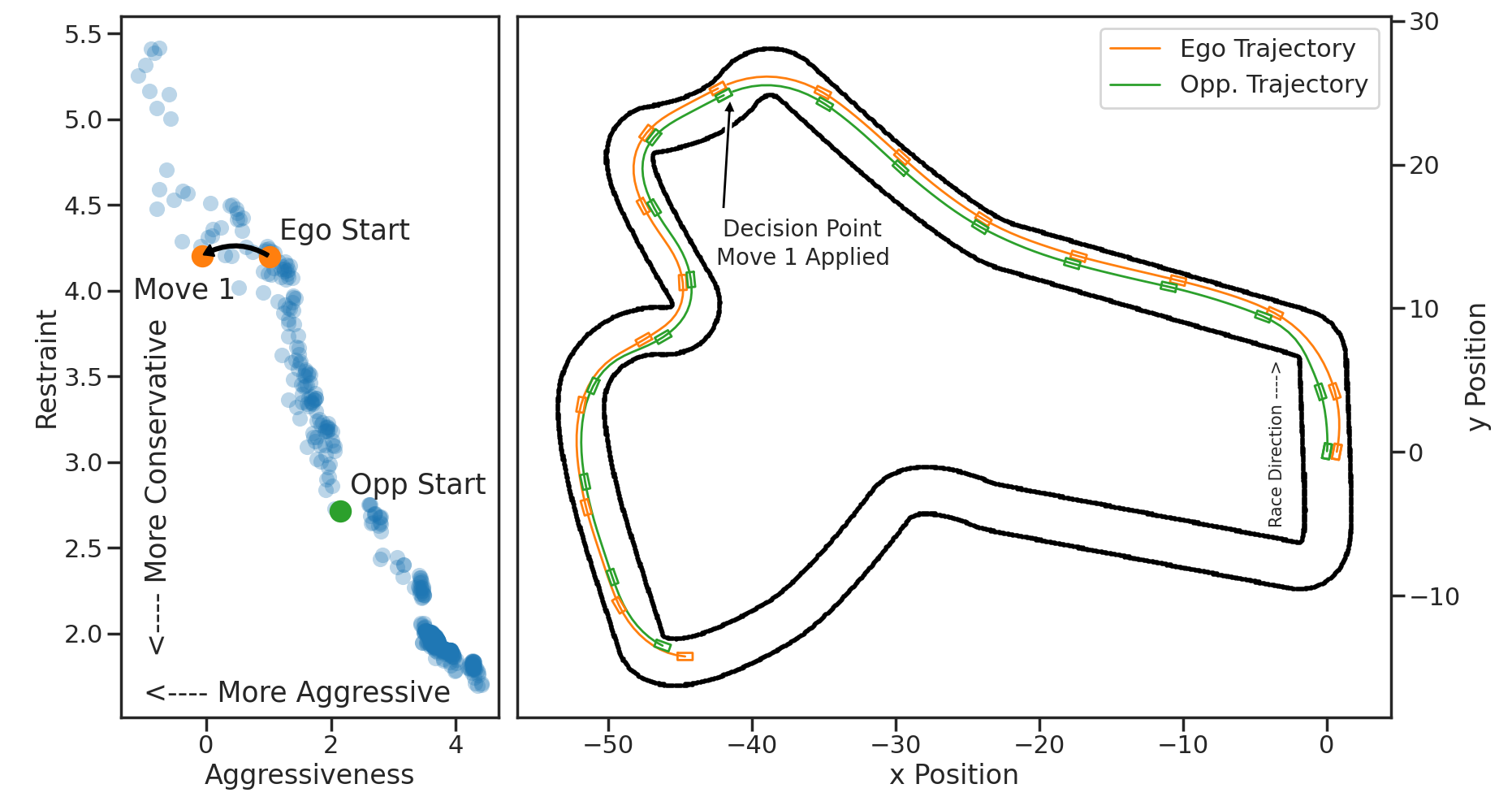}
    \caption{Effect of making a move in $\mathcal{O}$ in motion planning space.}
    \label{fig:cases_one}
\end{figure}
In this section, we examine specific cases of agents moving in the lower dimension and investigate whether we achieve the task of instilling explainability in our algorithm. We choose a 2-D space that encodes aggressiveness and restraint.
In Figure \ref{fig:cases_all}, we show four examples of races between two agents. The Ego (orange) uses our proposed game-theoretic approach in the Objective Space, and the Opponent (green) is static in the objective space. In the first two cases, the opponent is in the lower right quadrant, meaning that they're more conservative than aggressive. Hence our planner chooses to increase in aggressiveness continuously to win the races. In the last two cases, the opponent is in the upper left quadrant, meaning that they're more aggressive than conservative. Hence our planner chooses to first become more conservative, and once a chance presents itself, increase in aggressiveness and win the races.
In Figure \ref{fig:cases_one}, we inspect a specific move to show the effect in the motion planning space. In the beginning of the rollout, both agents are side by side. At a decision point, our planner locates the opponent in the lower right quadrant as more conservative than aggressive. Then the ego decides to increase in aggressiveness to finally overtake the opponent.
From these examples, it is clear that moving the planning problem in a lower dimension that encodes interpretable metrics for humans doesn't decrease the capability of the planner, or the completeness in the algorithm.
\section{Explainability in Control}
With robust localization, the autonomous vehicle can plan its trajectory but requires a safe an interpretable controller to execute the plan. In this section, we show our recent progress on learning-based safety-critical control through control barrier functions (CBFs) and provide our positioning on explainable safe control.
Specifically, we show that our proposed differentiable safety filter has more completeness than non-differentiable safety filter without sacrificing interpretability.

\subsection{Safety-critical control}
Learning-based control could provide high empirical performance thus it has become  popular for controlling complex dynamical systems.
However, learning-based controllers, such as neural network (NN), generally lack formal safety guarantees because of their black-box nature. This limits their deployments with complex safety-critical systems.
To address safety risk, multiple methods have been proposed, such as model predictive control (MPC)~\cite{camacho2013model}, Hamilton-Jacobi reachability analysis~\cite{bansal2017hamilton}, contraction theory~\cite{chou2022safe}, and control barrier functions (CBF)~\cite{ames2016control}.

Among the many safety-critical control techniques, CBF is becoming a popular choice since it explicitly specifies a safe control set by using a Lyapunov-like condition and guards the system inside a safe invariant set. When a continuous-time control-affine system is considered, such projection reduces to a convex quadratic program (QP) which is referred to as CBF-QP. Due to its simplicity, flexibility, and formal safety guarantees, CBFs have been applied in safe learning control with many successful applications~\cite{anand2021safe, dawson2022safe, taylor2020learning, dean2020guaranteeing, cheng2019end}.

\begin{figure}
    \centering
    \includegraphics[width = 0.23\textwidth]{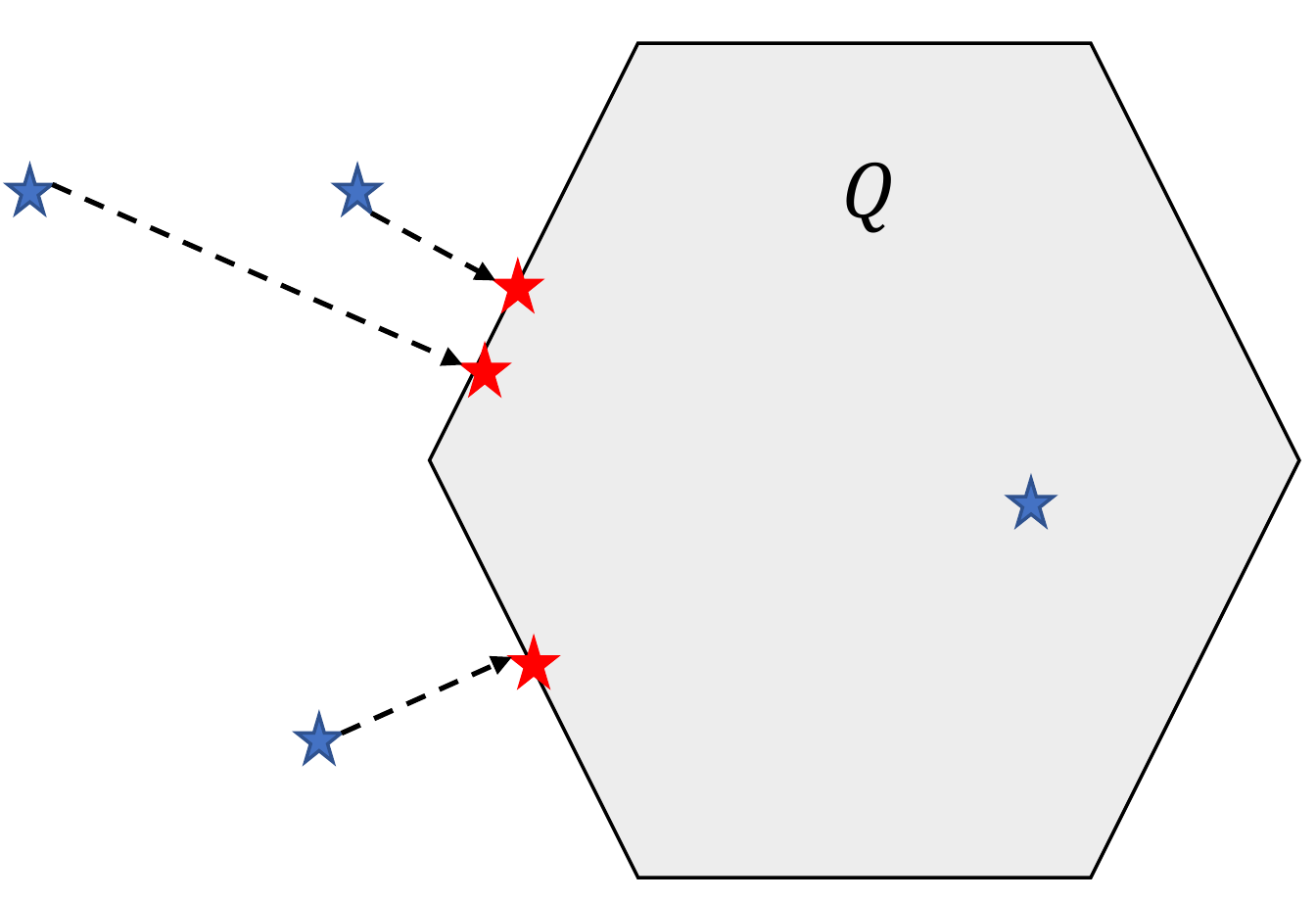}
		\caption{Illustration of QP: If the NN output is not in the safe control set $Q$, it will be projected in minimally invasive way to $Q$; otherwise, the control keeps the same.}
    \label{fig:qp_illustration}
\end{figure}

Compared with MPC, which needs to handle a possibly nonconvex optimization problem in the face of nonlinear dynamical systems, CBF-QP is computationally efficient to solve online. However, unlike MPC, the QP-based safety filter only operates in a minimally invasive manner, i.e., it generates the safe control input closest to the reference control input (in the Euclidean norm), as shown in Fig.~\ref{fig:qp_illustration}, unaware of the long-term effects of its action. This indicates that the effects of the safety filter on the performance of the closed-loop system are hard to predict. Therefore, the application of the safety filter may give rise to myopic controllers~\cite{cohen2020approximate} that induce subpar performance in the long term.

\begin{figure}
    \centering
    \includegraphics[width = 0.37\textwidth]{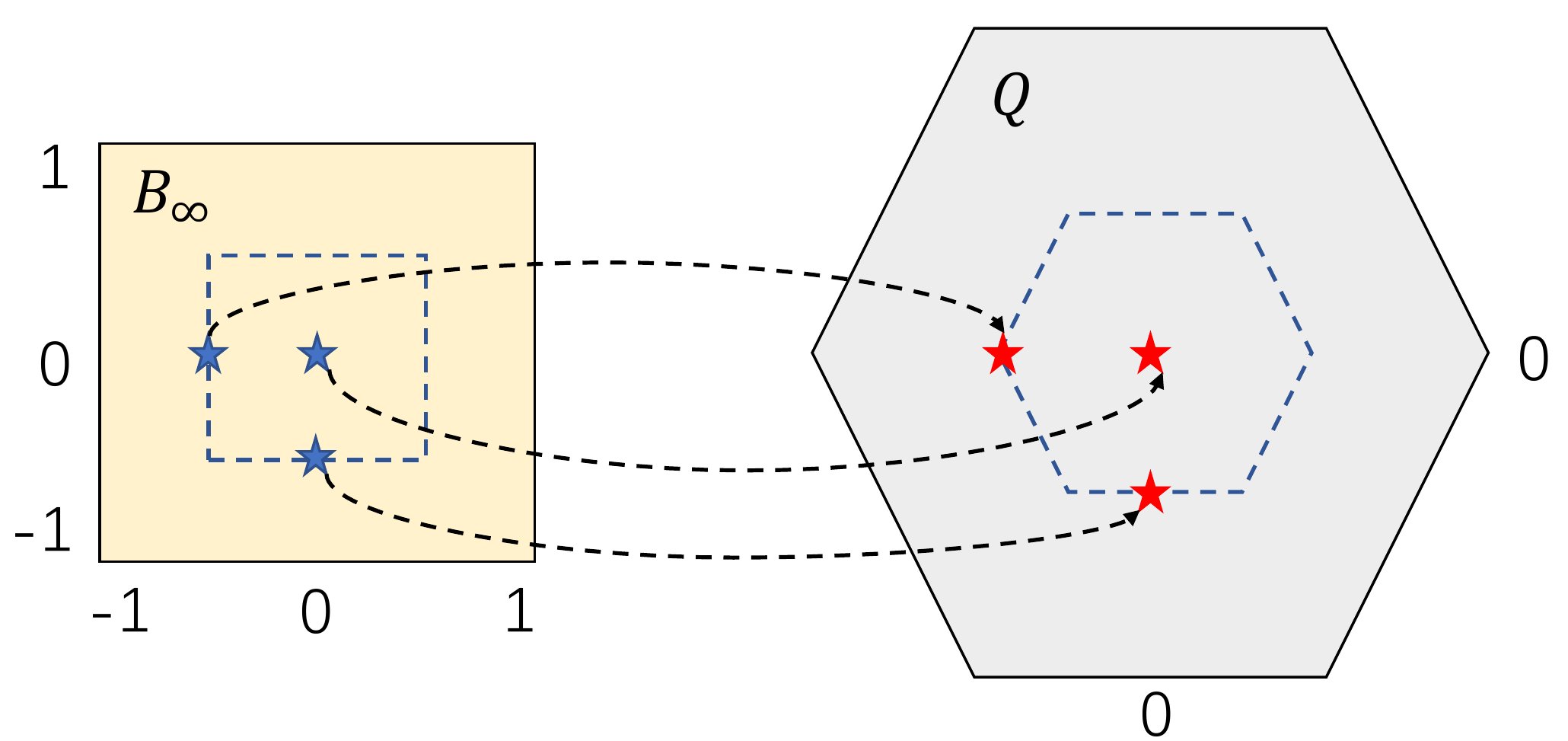}
		\caption{Illustration of the gauge map from the $\ell_\infty$ ball $B_\infty$ to a polytopic set $Q$. The original point in $B_\infty$ and mapped point in $Q$ are in the same level set and with the same direction.}
\label{fig:gauge_illustration}
\end{figure}

To address the issue of myopic CBF-based safety filters, in our recent work~\cite{yang2022differentiable}, we propose to utilize CBF to construct safe-by-construction NN controllers that allow end-to-end learning.
Incorporating safety layers in the NN controller allows the learning agent to take the effects of safety filters into account during training in order to maximize long-term performance.

\begin{figure}
    \centering
    \includegraphics[width = 0.4\textwidth]{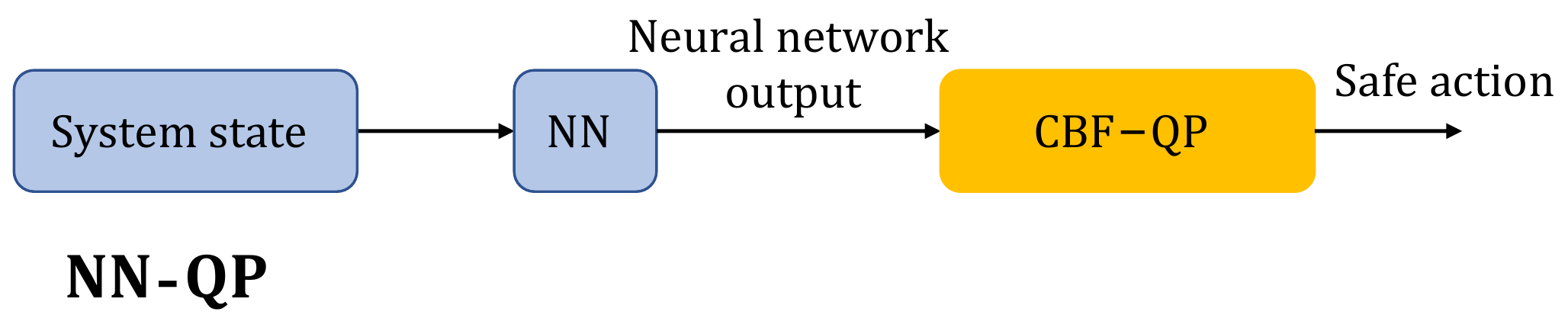}
    \caption{NN controller with CBF-QP safety filter.}
    \label{fig:qp}
\end{figure}

\begin{figure}[!t]
	\begin{subfigure}{0.98 \columnwidth}
		\centering 
		\includegraphics[width = \textwidth]{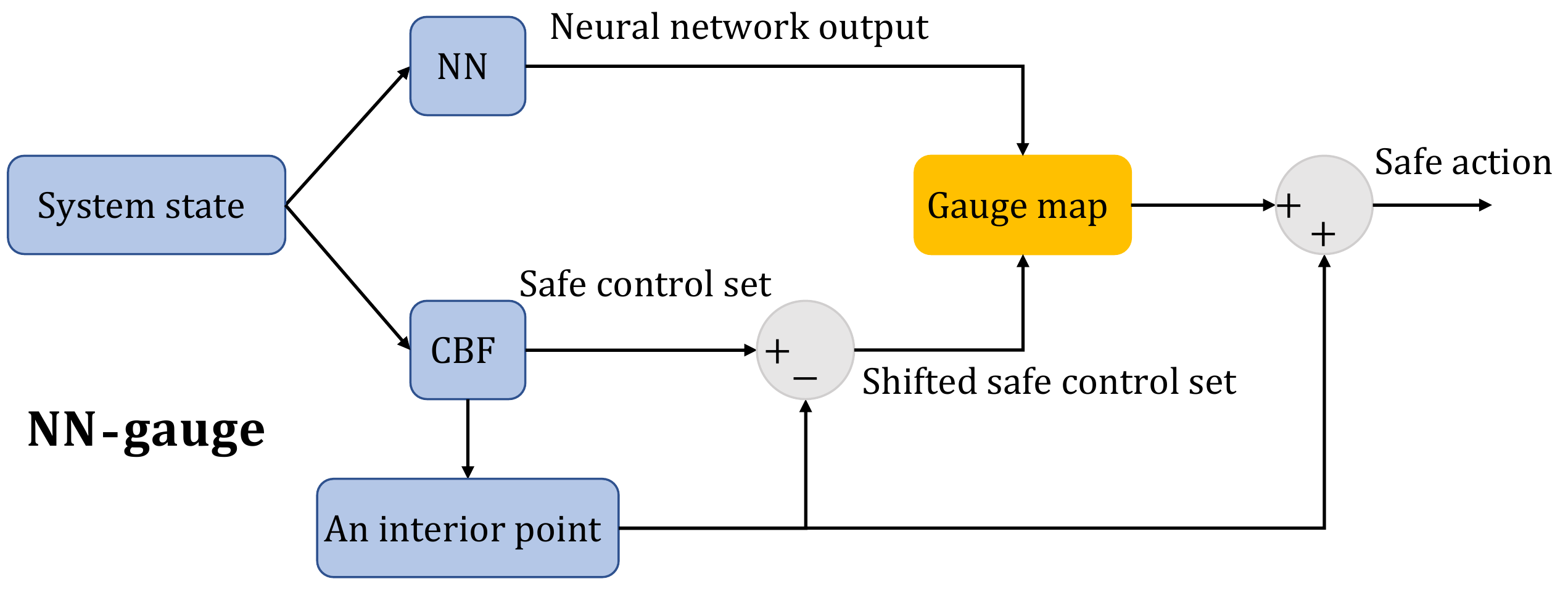}
		\caption{Gauge map-based safe NN controller architecture.}
  \label{fig:gauge_nn}
	\end{subfigure}
\hfil
	\begin{subfigure}{0.93 \columnwidth}
		\centering
		\includegraphics[width = \textwidth]{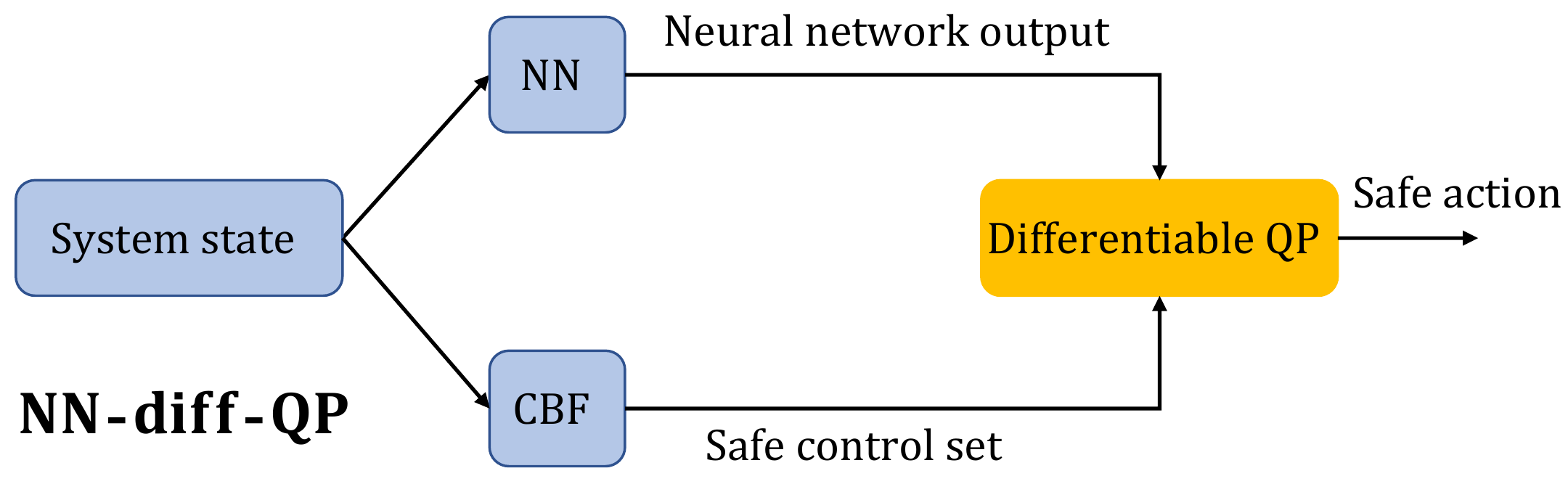}
		\caption{CBF-QP-based safe NN controller architecture.}
  \label{fig:diff_qp_framework}
	\end{subfigure}
\caption{Safe-by-construction NN controllers that utilize CBFs to construct differentiable safety layers (yellow blocks). }
\vspace{-10pt}
\label{fig:architecture}
\end{figure}

We design a differentiable safety layer using the gauge map~\cite{tabas2022computationally} (as shown in Fig.~\ref{fig:gauge_illustration}) which establishes a bijection mapping between the polytopic set of a NN output (e.g., an $\ell_\infty$ norm ball) and the CBF-based safe control set. The proposed architecture is denoted as NN-gauge (Fig.~\ref{fig:gauge_nn}). We compare NN-gauge with an alternative differentiable safe-by-construction NN controller called NN-diff-QP, which consists of a NN followed by a differentiable CBF-QP layer (Fig.~\ref{fig:diff_qp_framework}). Specifically, NN-diff-QP (Fig.~\ref{fig:diff_qp_framework}) concatenates a differentiable projection layer which can be implemented in a NN using toolboxes such as cvxpylayers~\cite{agrawal2019differentiable}, qpth~\cite{amos2017optnet}. 
NN-gauge involves finding an interior point of the safe control set since gauge map requires that $Q$ set in Fig.~\ref{fig:gauge_illustration} is convex and cover the origin. It can be reduced to implicitly finding the Chebyshev center~\cite{boyd2004convex} of the safe control set.

\begin{remark}
In the online execution, NN-gauge requires closed-form evaluation or solving a linear program (LP) while NN-diff-QP solves a quadratic program. Both methods are significantly cheaper to run than MPC. 
\end{remark}

\begin{figure}[!t]
\centering
	\begin{subfigure}{0.49 \columnwidth}
    \centering
    \includegraphics[width=\textwidth]{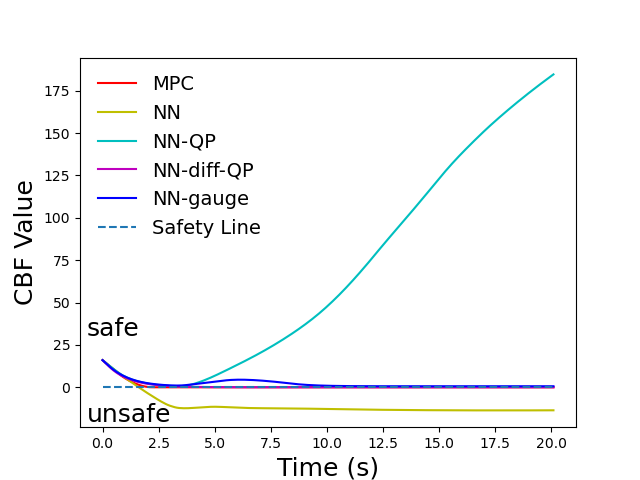}
    \caption{CBF values.}
    \label{fig:accCBFvalue}
	\end{subfigure}
		\begin{subfigure}{0.49 \columnwidth}
    \centering
    \includegraphics[width=\textwidth]{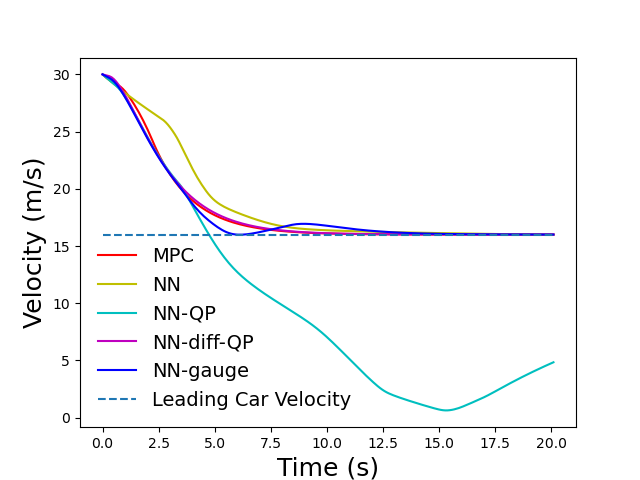}
    \caption{Velocity of the ego car.}
    \label{fig:accVelocity}
	\end{subfigure}
\caption{Results of adaptive cruise control. CBF values (left) and velocity of the ego car (right) under different controllers are evaluated in closed-loop for $20$s. A CBF value below zero indicates unsafety, and the optimal behavior of the ego car is expected to have a steady state velocity of $16 m/s$, same as the leading car. }
\label{fig:acc}
\end{figure}

As an example, let's consider the adaptive cruise control (ACC) in which the ego car is expected to achieve the desired cruising speed while maintaining a safe distance from the leading car. As shown in Fig.~\ref{fig:accVelocity}, applying the CBF-QP safety filter directly to enforce safe control deteriorates the long-term closed-loop performance of the NN controller. However, both NN-gauge and NN-diff-QP achieve similar closed-loop performance, and they are comparable to MPC.

\subsection{Explainability of safety-critical control}
\subsubsection{Non-differentiable QP filter}
Quadratic program (QP) is the classic minimally invasive safety filter to modify the nominal unsafe controller. Due to its simplicity and the nature of minimal modification of the original controller, it has always been equipped to CBF as the safety filter.
As shown in Fig.~\ref{fig:qp_illustration}, it is very understandable to humans, since one only needs to project the unsafe control input to the ``nearest'' point in the safe control set, which is specified by CBF.
Thus, QP safety filter enjoys high interpretability.

However, on the other hand, the QP operation is not necessarily the best choice from the perspective of optimizing a long-term system objective.
For example, in the long run, sometimes it might be better to choose a ``farther'' point in the safe control set than the ``nearest'' point on the boundary when the nominal control is not safe.
These modified actions are unaware of the long-term effects and may give rise to myopic controllers which deteriorate their performance.
From the ACC case study, we can also observe in Fig.~\ref{fig:accVelocity} that the modified control (i.e., NN-QP) is myopic with unpredictable long-term effects.
Thus, vanilla QP filter lacks the completeness in the sense of controlling the system in the optimal way with the safety guarantee.

\subsubsection{Differentiable safety filters}
Our proposed two differentiable safety filters, i.e., differentiable QP and gauge map, enjoy both a high level of interpretability and completeness.
\begin{itemize}
    \item \textbf{Differentiable QP}: the QP structure is the same with vanilla CBF-QP, so there is no interpretability loss in differentiable QP. On the other hand, since the NN controller considers the gradient of QP structure while training to minimize the long-term loss, so the trained NN adapts to QP much better than non-differentiable QP. Thus, even if the QP is applied to ensure safety when evaluating online, the modified controller could still have good performance. Therefore, from the explainability perspective, completeness is improved with no loss on interpretability.
    \item \textbf{Gauge map}: unlike QP, gauge map seeks to map the original control ``proportianally'' to the safe control set with the same direction. This mapping is also understandable to humans. Furthermore, since gauge map is sub-differentiable itself, so the end-to-end training holds naturally, which allows it to perform as good as differentiable QP after training. Thus, explanation could find the high completeness in this method as well.
\end{itemize}
In the ACC example, it also has been shown that both NN-gauge and NN-diff-QP have comparable performance with MPC, which implies the high completeness from demonstration. Looking ahead, we are interested in exploring a more general parameterized gauge map method, which even not necessarily maps to the same level set. In that case, it will perhaps lose some interpretability but have better completeness as it is not limited to the same level set mapping. This way, there is more flexibility while choosing the safe control input.
\section{Discussion and Conclusion}
In any autonomous vehicle which operates a sense-plan-act processing loop, the essential components are localization, planning and control. For humans to trust the system to make decisions and actions on their behalf in safety-critical situations, it is essential to have explainability across the stack. 
We view explainability as a trade-off between \emph{interpretability} of the machine's actions and the \emph{completeness} in terms of describing the operation of the system in an accurate way. 

The explainability of localization methods generally lies in the uncertainty estimation and the ability to explain and tune the methods when localization fails. We introduce Local\_INN, which utilizes an invertable neural network architecture that provides explainability for uncertainty estimation and explainability from map representation. 
Uncertainty estimation, through the covariance of the inferred pose distributions, improves explainability by providing information on the measurement quality of the prediction. Furthermore, the guaranteed invertibility of the neural network provides the use a direct way to check the neural network’s understanding of
the map by analyzing the reconstructed map. This feedback from the internals of the localization engine allows the human to know where in the map the network has lower confidence and that they should augment it with more data.

The expainability in planning is necessary to ensure the autonomous vehicle maintains a safe trajectory at all times in terms of  maximizing the distance of a planned trajectory to obstacles and undrivable areas,
maximizing velocity on the trajectories, minimizing the lateral acceleration the vehicle experiences on a planned trajectory, and so on. To achieve this, we introduced a new approach on Game-Theoretic Objective Space Planning where we map these complex planning objectives to a lower dimensional space of balancing aggressiveness and restraint. In a racing context, we show how our planner increases aggressiveness continuously to win the races. Similarly, it chooses to be more conservative to maintain safety. By moving the planning problem
in a lower dimension that encodes interpretable metrics for humans we demonstrate how it doesn’t decrease the capability of the planner, or reduce the
completeness in the algorithm.

Finally, to ensure the plan is executed in a safe manner, we describe our efforts in explainable safe neural network controller design with Differentiable CBF-QP filters. The Control Barrier Function structure of the filter ensures the control output is always safe in an interpretable manner as there is a direct projection of the NN controller output into the safe control set. The performance of the proposed safe controllers is comparable to model-predictive controllers, which implies the high completeness from demonstrations. 

Through these three learning-based architectures for localization, planning and safe control, we have demonstrated the initial findings on explainability of decisions in the sense-plan-act process of autonomous systems. Further developments in explainability of autonomous systems will lead to more public trust and safer systems.

\bibliographystyle{IEEEtran}
\bibliography{main.bib}

\end{document}